\pdfoutput=1

\documentclass[11pt]{article}

\usepackage{acl}

\usepackage{times}
\usepackage{latexsym}

\usepackage[T1]{fontenc}

\usepackage[utf8]{inputenc}
\usepackage{CJKutf8}
\usepackage{graphicx}
\usepackage{svg}
\usepackage{amsmath}
\usepackage{amssymb}
\usepackage{multicol,multirow}
\usepackage{xcolor}
\usepackage{booktabs}

\usepackage[T5]{fontenc}        
\usepackage{url}

\usepackage{microtype}

\usepackage{inconsolata}

\usepackage{graphicx}
\usepackage{subfigure}

\title{XCOMPS: A Multilingual Benchmark of Conceptual Minimal Pairs}

\author{Linyang He\thanks{~~Equal contribution.}\textsuperscript{1}\quad Ercong Nie\footnotemark[1]\textsuperscript{2, 3}\\
\textbf{Sukru Samet Dindar\textsuperscript{1}\quad Arsalan Firoozi\textsuperscript{1}\quad Adrian Florea\textsuperscript{1}\quad}\\
\textbf{Van Nguyen\textsuperscript{3}\quad Corentin Puffay\textsuperscript{5}\quad Riki Shimizu\textsuperscript{1} \quad Haotian Ye\textsuperscript{2,3} } \\ 
\textbf{Jonathan Brennan\textsuperscript{4} \quad Helmut Schmid\textsuperscript{3} \quad Hinrich Sch\"utze\textsuperscript{2, 3}\footnotemark[2]\quad Nima Mesgarani\textsuperscript{1}\thanks{~~Corresponding authors.}}\\
\textsuperscript{1}Columbia University ~\textsuperscript{2}Munich Center for Machine Learning\\
\textsuperscript{3}LMU Munich ~\textsuperscript{4}University of Michigan ~\textsuperscript{5}KU Leuven\\
 \texttt{linyang.he@columbia.edu} \quad \texttt{nie@cis.lmu.de} \quad  \\
\texttt{hinrich@hotmail.com}\quad  \texttt{nima@ee.columbia.edu} \quad 
}

\begin{document}
\maketitle
\begin{abstract}
In this work, we introduce XCOMPS, a multilingual conceptual minimal pair dataset that covers 17 languages.
Using this dataset, we evaluate LLMs' multilingual conceptual understanding through metalinguistic prompting, direct probability measurement, and neurolinguistic probing. 
We find that: 1) LLMs exhibit weaker conceptual understanding for low-resource languages, and accuracy varies across languages despite being tested on the same concept sets.  2) LLMs excel at distinguishing concept-property pairs that are visibly different but exhibit a marked performance drop when negative pairs share subtle semantic similarities.
3) More morphologically complex languages yield lower concept understanding scores and require deeper layers for conceptual reasoning.
The dataset is publicly available at: \url{https://github.com/LinyangHe/XCOMPS/}.
\end{abstract}

\section{Introduction}
Large language models (LLMs) have demonstrated remarkable capabilities across various natural language understanding (NLU) tasks.
Recent advances, such as GPT-4 \cite{achiam2023gpt} and Llama 3 \cite{dubey2024llama}, have shown that LLMs can produce human-like outputs and handle complex linguistic phenomena. 
However, 
whether LLMs genuinely understand semantics or merely rely on shallow statistical correlations is disputable \cite{lake2018generalization,elazar2021measuring,huang2023survey}. 
One fundamental aspect of human conceptual understanding is that it is not dependent on specific linguistic forms or modalities \cite{carey2000origin,mandler2004foundations}. When humans learn and reason about concepts, they do not require the knowledge to be tied to a particular medium, such as text, images, or video, nor do they rely on a specific language. 
This raises an important question: \textit{Does LLMs' conceptual-property reasoning remain stable across languages, or is it language-specific?}


To explore this, \citet{misra2023comps} introduced the COMPS dataset, designed to probe the semantic reasoning abilities of LLMs through minimal pairs in English. However, COMPS only evaluates monolingual conceptual-property reasoning, leaving open the question of whether LLMs generalize such reasoning across languages. In this work, we introduce XCOMPS, a multilingual extension of COMPS, to assess whether LLMs' semantic reasoning is universally consistent across languages. XCOMPS covers 17 languages, including analytic, inflectional, and agglutinative languages, ensuring a broad representation of linguistic structures. 

\begin{table*}[t]
\flushleft
\scalebox{0.675}{
\begin{tabular}{llll}
\hline
\textbf{Type}          & \textbf{Language}   & \textbf{Acceptable Sentence}                                                                                                                   & \textbf{Unacceptable Sentence}                                                                                                             \\ \hline
Taxonomic     & Spanish    & \begin{tabular}[c]{@{}l@{}}\textit{Tostadora} se utiliza para calentar alimentos.\\ (A toaster is used for heating food.)\end{tabular}         & \begin{tabular}[c]{@{}l@{}}\textit{Cafetera} se utiliza para calentar alimentos.\\ (A coffee maker is used for heating food.)\end{tabular} \\
Overlap       & Vietnamese & \begin{tabular}[c]{@{}l@{}}\textit{Máy nướng bánh mì được} sử dụng để hâm nóng thực phẩm.\\ (A toaster is used for heating food.)\end{tabular} & \begin{tabular}[c]{@{}l@{}}\textit{Tủ lạnh được }sử dụng để hâm nóng thực phẩm.\\ (A refrigerator is used for heating food.)\end{tabular}  \\
Co-occurrence & Hungarian  & \begin{tabular}[c]{@{}l@{}}\textit{Kenyérpirító} ételek melegítésére használják.\\ (A toaster is used for heating food.)\end{tabular}          & \begin{tabular}[c]{@{}l@{}}\textit{Vízforraló} ételek melegítésére használják.\\ (A kettle is used for heating food.)\end{tabular}         \\
Random        & Dutch      & \begin{tabular}[c]{@{}l@{}}\textit{Broodrooster} wordt gebruikt om voedsel te verwarmen.\\ (A toaster is used for heating food.)\end{tabular}  & \begin{tabular}[c]{@{}l@{}}\textit{Winterkoning} wordt gebruikt om voedsel te verwarmen.\\ (A wren is used for heating food.)\end{tabular} \\ \hline
\end{tabular}}
\caption{\small XCOMPS examples, illustrating each linguistic variant pairs an acceptable sentence (positively matched property) with an unacceptable counterpart (negatively matched property).}
\label{tab:xcomps}
\end{table*}

Beyond dataset expansion, evaluating LLMs' reasoning abilities has increasingly relied on prompt engineering, often referred to as metalinguistic prompting \cite{hu2023prompting}. However, recent work \cite{hu2023prompting,he2024large} suggests that metalinguistic prompting primarily assesses performance---that is, how well a model produces correct outputs---rather than its underlying competence in conceptual understanding. This distinction is crucial, as models may perform well on explicit prompts but lack true conceptual representations \cite{piantadosi2022meaning}. 
To investigate LLMs' multilingual capabilities and determine whether they genuinely encode conceptual knowledge across languages, we adopt a three-pronged evaluation approach: \textit{Metalinguistic prompting}, \textit{Neurolinguistic probing}, and \textit{Direct probability measurement}.
Our experimental results reveal several insights into the multilingual conceptual reasoning capabilities of LLMs: 
1) Conceptual understanding is not consistently maintained across languages. Even when models perform well in English, their reasoning ability deteriorates significantly in low-resource languages; the extent of deterioration also varies across different low-resource languages. 2) Models perform well when conceptual relationships are highly distinct but struggle with subtle semantic distinctions. 
3) Languages with higher morphological complexity (agglutinative > inflected > analytic) yield lower concept-reasoning scores.
These results suggest that LLMs' semantic reasoning may not generalize universally across linguistic boundaries.

\section{Language Performance vs. Competence}
As suggested in \citet{he2024large}, LLMs can be evaluated through three methods: \textit{metalinguistic prompting}, which assesses \textit{performance} based on explicit responses; direct probability measurement, which provides an intermediate evaluation by comparing model-generated probabilities; and \textit{neurolinguistic probing}, which directly examines \textit{competence} by analyzing internal activation patterns\footnote{For simplicity, we refer to these three methods as Meta, Direct, Neuro.}.
\paragraph{Metalinguistic Prompting for Performance} This method involves explicitly querying the model about linguistic expressions, often in a comparative or multiple-choice format. By asking the model to choose between minimal pairs (e.g., ``Which sentence is more grammatically correct?''), researchers can evaluate how well the model retrieves and verbalizes knowledge. Using prompting, researchers have revealed new classes of emergent abilities such as arithmetic, instruction-following, grounded conceptual mappings, and sentence acceptability judgments \cite{brown2020language,wei2022chain,patel2021mapping,dentella2023testing}.
Because the responses are influenced by prompt engineering and surface-level cues, this method primarily reflects performance rather than deep conceptual competence.
\paragraph{Direct Probability Measurement} Instead of relying on explicit responses, this method examines the model's probability assignment to different sentences within minimal pairs. For example, a model should assign a higher probability to `A robin can fly' than to `A penguin can fly'. This approach offers a more objective evaluation than metalinguistic prompting and captures implicit model preferences, placing it between performance and competence.  Researchers have designed syntactic, semantic/conceptual, and discourse inference tasks using the probability assignment method, offering different insights into LLMs' capabilities compared to metalinguistic prompting \cite{futrell2019neural,gauthier2020syntaxgym,hu2020systematic,warstadt2020blimp,beyer2021incoherence,misra2023comps,kauf2023event}.
However, it still relies on external outputs and does not fully reveal how the model internally represents concepts.

\paragraph{Neurolinguistic Probing for Competence} This approach goes beyond external outputs by analyzing internal activation patterns across different layers of the model~\cite{he2024decoding,he2024large}. Using diagnostic classifiers, researchers can probe whether LLMs inherently encode conceptual-property relationships or simply rely on statistical correlations. Since it provides a direct measure of competence, neurolinguistic probing is more reliable for assessing the depth of linguistic understanding.

\section{XCOMPS}

\subsection{Concept Selection}
To ensure that XCOMPS maintains conceptual alignment with COMPS while extending its scope to multiple languages, we use the same 521 concepts and their negative samples from COMPS. 
As shown in Table \ref{tab:xcomps}, these negative samples can be categorized into three types. \textit{Taxonomy-based} negative samples are selected based on hierarchical relationships among concepts. 
Negative samples come from the same broad category as the positive concept but differ in key property attributions.
\textit{Property norm-based (overlap)} negative samples are chosen based on shared semantic properties with the positive concept while lacking the specific property under evaluation. 
\textit{Co-occurrence-based samples} are selected from concepts that frequently appear in similar contexts but do not share the target property. 
XCOMPS also has additional \textit{random negative concepts} from the set of concepts that do not possess the property of the original positive concept. 

\subsection{Properties of Concepts}
In XCOMPS, the properties assigned to concepts are inherited from COMPS, ensuring alignment across languages while maintaining the original conceptual-property relationships. These properties in COMPS were originally derived from the XCSLB dataset, an extended version of the CSLB property norm dataset \cite{devereux2014centre}, which captures human-annotated perceptual, functional, and categorical attributes of concepts. Additionally, taxonomic relationships from resources like WordNet \cite{miller1995wordnet} were used to infer properties through hierarchical inheritance, ensuring that general category attributes (e.g., ``mammals have fur'') are systematically applied to their subcategories. Some properties also reflect real-world associations observed in corpus-based co-occurrence statistics. 

\subsection{Multilingual Data Construction}
To construct XCOMPS, which covers 17 languages (Table \ref{lang_info} in Appendix \ref{appendix}), we adopted a human-LLM interactive translation pipeline, leveraging both human expertise and the multilingual generation capabilities of LLMs. The language set for XCOMPS aligns with the prior knowledge probing benchmarks, such as BMLAMA-17~\cite{qi-etal-2023-cross} and KLAR~\cite{wang2025lost}, ensuring consistency in multilingual evaluation. 
The highly structured nature of conceptual minimal pair datasets, where positive and negative sentences primarily consist of two components--concepts and properties--enabled us to design a multi-step translation process that ensures high-quality multilingual data.

The construction process consists of four stages. We use the GPT-4o model (\texttt{GPT-4o-2024-08-06}) via the OpenAI API as the translation assistant in the pipeline. 
In the first stage, we manually translated the original concepts and properties from English into German and Chinese using language experts. We used German and Chinese as additional seed languages to further reduce ambiguity, This multilingual seed data helped disambiguate concepts that might otherwise be unclear in translation. For example, the English word ``bat'' could refer to either the flying animal or the sports equipment. By including the German term ``Schl\"ager'' and the Chinese term ``\begin{CJK}{UTF8}{gbsn}球拍\end{CJK}'', which both unambiguously refer to the sports equipment, we ensured that the intended concept was accurately captured during translation.

In the second stage, we used LLMs to expand the seed data into the remaining 15 languages. LLMs were tasked with translating the concepts and properties, leveraging their multilingual machine translation capabilities. By providing seed data in three languages (English, German, and Chinese), we enhanced the LLMs' ability to generate accurate translations, as the additional context reduced the likelihood of semantic errors.

In the third stage, human experts for each target language manually reviewed and corrected the translated concepts and properties. This step ensured that the translations were accurate, culturally appropriate, and semantically aligned with the original dataset. Human intervention was particularly critical for low-resource languages, where LLMs often struggle with semantic precision in translation tasks.

Finally, in the fourth stage, LLMs were employed to generate complete sentences based on the verified concepts and properties. This step involved formulating positive and negative sentence pairs, which can be viewed as a straightforward language manipulation task. By providing the translated concepts and properties as input, we enabled the LLMs to focus on generating fluent and grammatically correct sentences, leveraging their strengths in multilingual text generation. This approach ensured that the most challenging aspect of the task--accurate translation of concepts and properties--was already resolved, allowing the LLMs to produce high-quality outputs.

By splitting the process into property translation and sentence generation, using multilingual seed data to reduce ambiguity, and combining human expertise with LLM capabilities, we ensured the quality and consistency of the XCOMPS dataset. This human-LLM interactive translation pipeline demonstrates how LLMs' multilingual understanding and generation capabilities can be effectively harnessed to construct high-quality multilingual benchmarks.

\section{Experiment Setup}
\subsection{Model}
We use \texttt{meta-llama/Llama-3.1-8B-Instruct} from Hugging Face in our experiment, which applies instruction tuning to the base model for more intuitive user-prompt handling. During the inference, we adopt \texttt{float16} precision to minimize computational resource consumption while maintaining model performance.

\subsection{Evaluation}
For \textbf{Meta}, we present both sentences of a minimal pair within a single prompt. We convert the target property into a question and compare the probabilities assigned to acceptable vs. unacceptable concepts. 
Figure \ref{prompt_template} in Appendix \ref{appendix} shows the prompts used in the experiment.
For \textbf{Direct}, we compute sentence probabilities directly from the model's logits. A prediction is considered correct if the model assigns a higher probability to the valid sentence within each minimal pair. 
For \textbf{Neuro}, we adopt last-token pooling to represent each sentence, extracting the final token's hidden state from every layer. This approach ensures coverage of all preceding tokens \cite{SFRAIResearch2024}. We then apply a logistic regression classifier for probing, using the F1 score (averaged over five cross-validation folds) as our primary evaluation metric.


\begin{figure}[t]
    \centering

    \includegraphics[width=\linewidth]{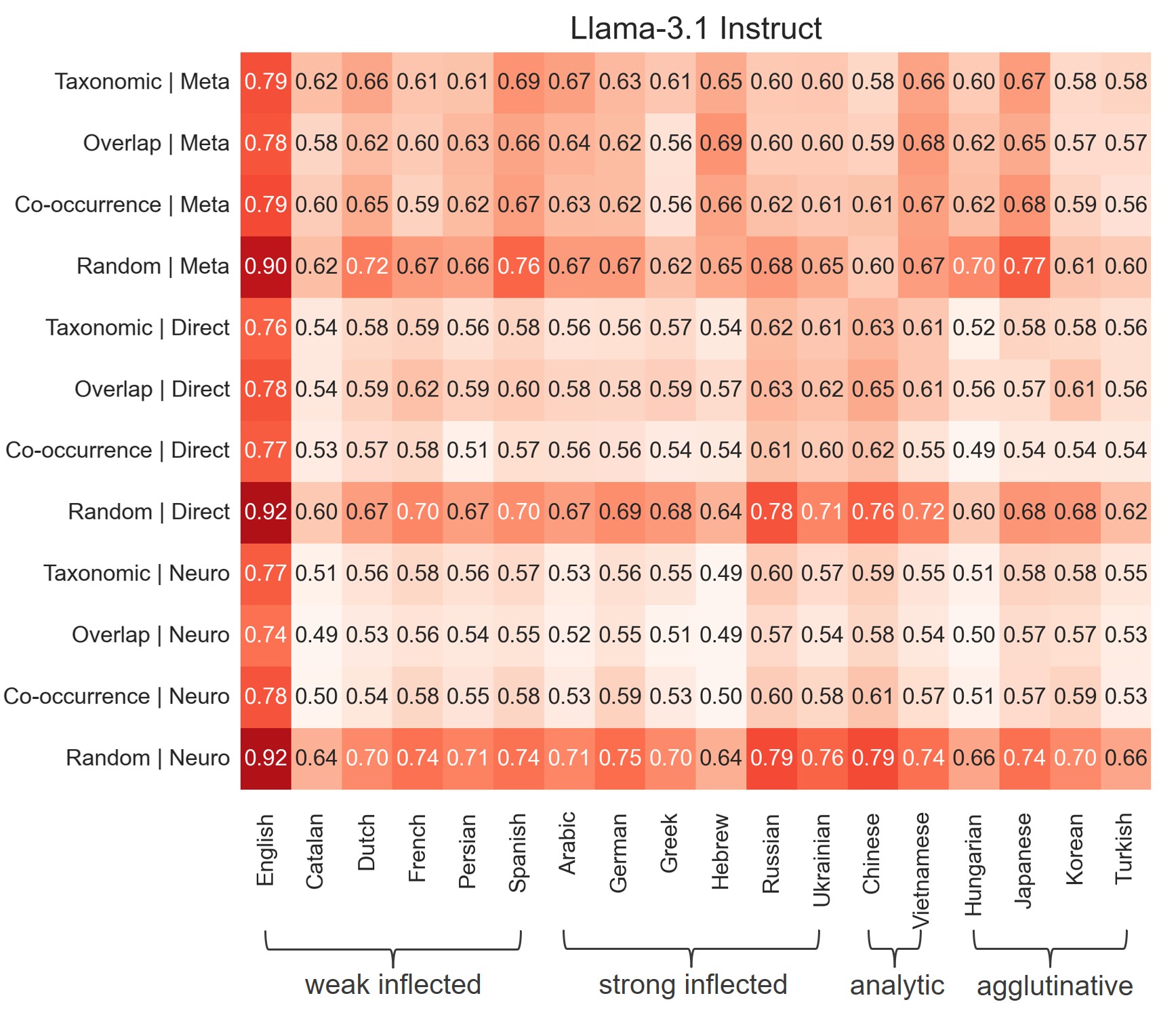}

    \caption{\small Metalinguistic prompting (meta), direct probability measurement (direct), and minimal pair probing (neuro) results on XCOMPS. The meta method evaluates LLMs' language performance; the neuro method evaluates LLMs' language competence, and the direct method falls in between. Languages are grouped according to morphological typology. Neuro-probing is a layer-wise method, and here we use the max value across all layers to compare with Meta and Direct.}
    \label{fig:mcomps_all}
\end{figure}

\subsection{Results and Analysis}
\paragraph{Cross-linguistic variability in conceptual reasoning.} From Figure \ref{fig:mcomps_all}, we observe that the model can perform relatively well on English conceptual tasks but show marked declines for low-resource languages. Notably, some languages with limited training data (e.g., Hungarian, Catalan) exhibit greater deterioration than others, indicating that cross-linguistic generalization of conceptual understanding is far from uniform. Even within the low-resource category, the degree of performance drop varies, underscoring that LLMs' semantic reasoning is neither universally stable nor equally supported by existing multilingual corpora. These patterns reinforce the idea that conceptual capabilities learned in English do not necessarily transfer seamlessly to languages that differ typologically or have weaker representations in training data.


\paragraph{Models excel at distinct conceptual contrasts but falter with subtler differences.} High scores all appear in Random rows, where the negative concept is clearly distinct (e.g., ``toaster'' vs. ``wren''), and the model easily detects mismatches. In Taxonomic, Overlap, or Co-occurrence rows, however, performance drops because the negative concepts share subtle semantic similarities (e.g., ``toaster'' vs. ``coffee maker''). This indicates that the models may rely on conspicuous cues rather than true conceptual reasoning.

\paragraph{Direct and neuro convergence.} By comparing direct and neuro results in Figure \ref{fig:mcomps_all}, and from Figure \ref{fig:linear_cor} in Appendix \ref{appendix}, we see high correlations across all negative types, indicating that direct measurements closely track the models' internal representations. 

\paragraph{Higher morphological complexity, lower conceptual reasoning.} Figure \ref{fig:mcomps_avg} in Appendix \ref{appendix} shows that languages with greater morphological complexity (moving from Analytic to Inflected to Agglutinative) tend to yield lower concept-reasoning scores. This indicates that, as linguistic structure becomes more complex, it becomes harder for the models to capture concept-property relationships consistently.

\section{Conclusion}
In this work, we introduce the XCOMPS benchmark, which provides a multilingual conceptual minimal pair dataset for evaluating the language model's semantic understanding across 17 languages. This work reveals that while LLMs demonstrate surface-level multilingual capabilities, they lack a universal semantic reasoning mechanism that transcends language boundaries. 

\section*{Limitation}
While XCOMPS significantly advances the evaluation of multilingual conceptual understanding, certain limitations remain. First, although the dataset covers 17 typologically diverse languages, it does not encompass all linguistic families or low-resource languages, which may limit its generalizability to underrepresented languages. Second, the reliance on human-LLM interaction for data construction ensures high quality but introduces potential inconsistencies due to variations in human expertise and model outputs. Lastly, while XCOMPS focuses on conceptual understanding, it does not explicitly address other challenges in multilingual NLP, such as pragmatics or contextual reasoning. Despite these limitations, XCOMPS provides a robust foundation for assessing and improving LLMs’ multilingual capabilities, and future work can extend its scope to address these areas.

\section*{Acknowledgement}
We thank the anonymous reviewers for their valuable advice and feedback.
This research was partially supported by DFG (German Research Foundation) grant SCHU 2246/14-1 and Munich Center for Machine Learning (MCML).

%
%
%
\bibliography{references}

\appendix

\section{Appendix}
\label{appendix}

\begin{figure*}
    \centering
\includegraphics[width=\linewidth]{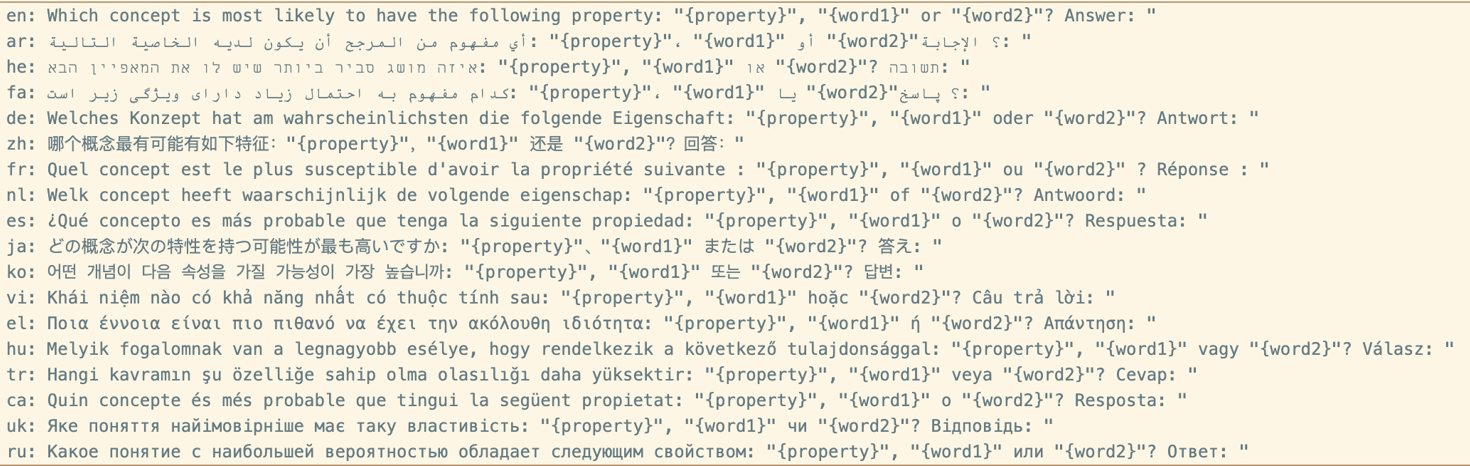}
    \caption{Prompt templates of different languages used for metalinguistic prompting.}
    \label{prompt_template}
\end{figure*}

\begin{figure*}[t]
    \centering
    \includegraphics[width=.8\linewidth]{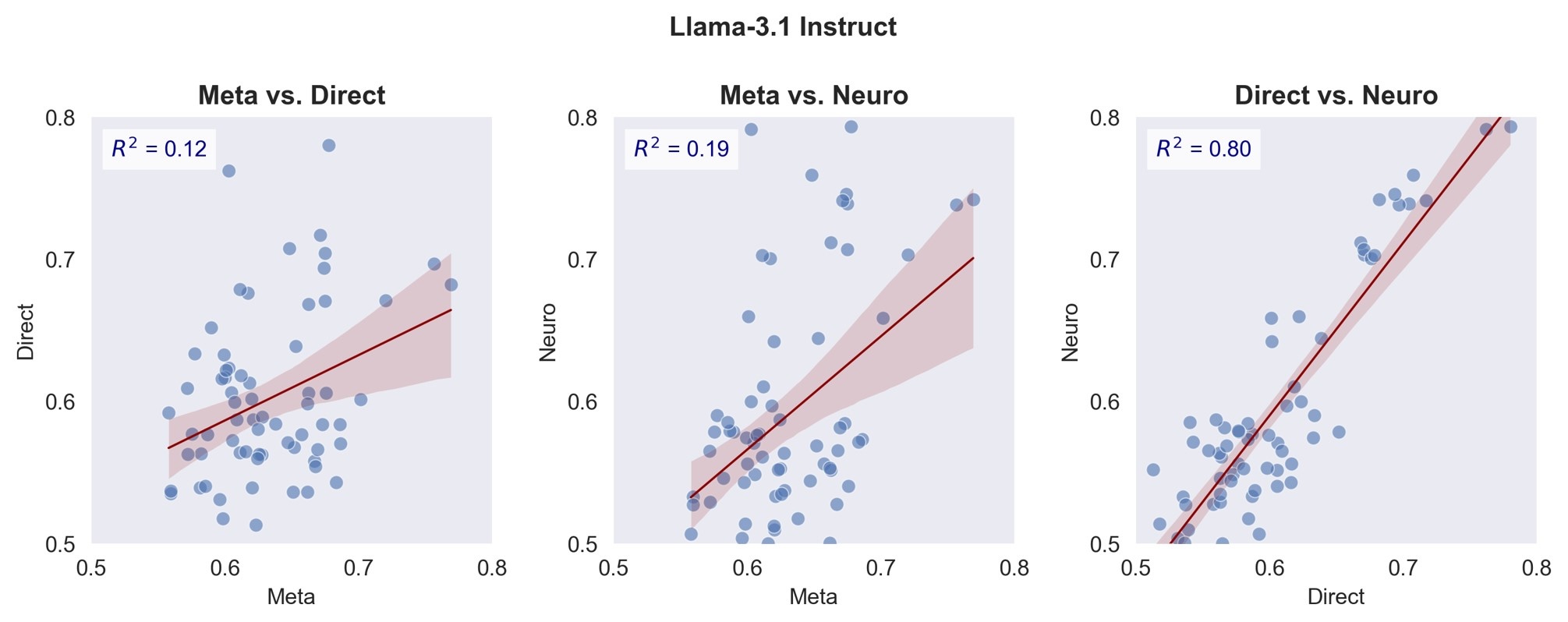}
    \caption{\small Linear correlation among meta, direct, and neuro evaluation results for all four tasks.}
    \label{fig:linear_cor}
\end{figure*}

\begin{figure}[t]
    \centering
    \includegraphics[width=.8\linewidth]{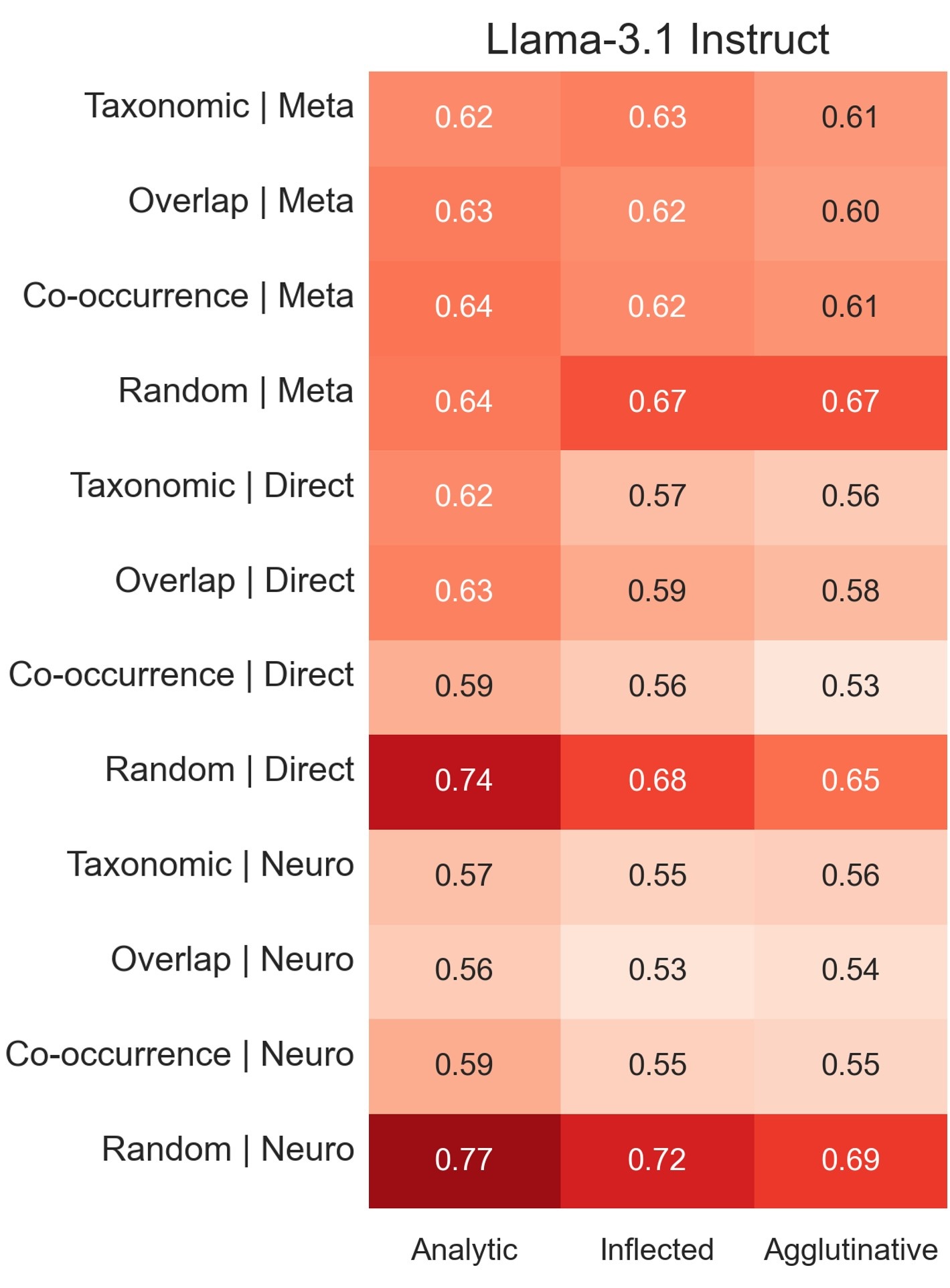}
    \caption{\small Averaged results across different language types. English results are dropped to make the comparison more reliable among low-resource languages.}
    \label{fig:mcomps_avg}
\end{figure}

Table \ref{lang_info} shows the detailed information of the languages covered by XCMOPS.
Figure \ref{prompt_template} displays the prompt templates of different languages used for metalinguistic prompting evaluation.
Figures \ref{fig:linear_cor} and \ref{fig:mcomps_avg} show detailed experimental results. 

\begin{table}[h]
\centering
\scalebox{.7}{
\begin{tabular}{clll} 
\toprule
\textbf{lid}   & \textbf{language} &  \textbf{Typology}  & \textbf{Family}   \\ 
\midrule

ar    & Arabic  & Inflectional    & Semitic     \\ 

ca    & Catalan  & Inflectional    & Indo-European (Romance)        \\ 

de    & German  & Inflectional     & Indo-European (Germanic)  \\ 

el    & Greek    & Inflectional    & Indo-European (Hellenic)    \\ 

es    & Spanish  & Inflectional    & Indo-European (Romance)     \\ 

fa    & Persian & Inflectional     & Indo-European (Iranian)     \\ 

fr    & French  & Inflectional     & Indo-European (Romance)      \\ 

he    & Hebrew  & Inflectional     & Semitic     \\ 
hu    & Hungarian &  Agglutinative    & Uralic      \\ 

ja    & Japanese &  Agglutinative     & Isolate     \\ 

ko    & Korean &  Agglutinative       & Isolate     \\ 

nl    & Dutch   & Inflectional     & Indo-European (Germanic)   \\ 

ru    & Russian   & Inflectional   & Indo-European (Slavic)      \\ 

tr    & Turkish &  Agglutinative  & Turkic      \\ 

uk    & Ukrainian & Inflectional   & Indo-European (Slavic)      \\ 

vi    & Vietnamese & Analytic & Austroasiatic    \\ 

zh & Chinese  & Analytic   & Sino-Tibetan      \\
\bottomrule
\end{tabular}}
\caption{Detailed information of the languages covered by XCOMPS.}
\label{lang_info}
\end{table}

\end{document}